\def\year{2019}\relax
\documentclass[letterpaper]{article} 
\usepackage{aaai19}  
\usepackage{times}  
\usepackage{helvet} 
\usepackage{courier}  
\usepackage[hyphens]{url}  
\usepackage{graphicx} 
\usepackage{graphicx}  
\usepackage{cuted}
\usepackage{graphicx}
\usepackage{subcaption}
\usepackage{amsmath}
\title{Assessing Data Quality of Annotations With Krippendorff's Alpha For Applications in Computer Vision}

\author{Joseph Nassar, Viveca Pavon-Harr, Marc Bosch, and Ian McCulloh\\ 
Accenture Federal Services\\
800 N. Glebe Rd.\\
Arlignton, VA\\
marc.bosch.ruiz@accenturefederal.com 
}

\begin{document}

\maketitle

\begin{abstract}
Current supervised deep learning frameworks rely on annotated data for modeling the underlying data distribution of a given task. In particular for computer vision algorithms powered by deep learning, the quality of annotated data is the most critical factor in achieving the desired algorithm performance. Data annotation is, typically, a manual process where the annotator follows guidelines and operates in a ``best-guess'' manner. Labeling criteria among annotators can show discrepancies in labeling results. This may impact the algorithm inference performance. Given the popularity and widespread use of deep learning among computer vision, more and more custom datasets are needed to train neural networks to tackle different kinds of tasks. Unfortunately, there is no full understanding of the factors that affect annotated data quality, and how it translates into algorithm performance. In this paper we studied this problem for object detection and recognition. We conducted several data annotation experiments to measure inter annotator agreement and consistency, as well as how the selection of ground truth impacts the perceived algorithm performance. We propose a methodology to monitor the quality of annotations during the labeling of images and how it can be used to measure performance. We also show that neglecting to monitor the annotation process can result in significant loss in algorithm precision. Through these experiments, we observe that knowledge of the labeling process, training data, and ground truth data used for algorithm evaluation are fundamental components to accurately assess trustworthiness of an AI system.
\end{abstract}

\section{Introduction}
With the rise of deep learning within computer vision, the spotlight is shifting from algorithm design to dataset development. Data is the highest contributor to model performance for many modern neural network architectures. Adding layers to the network, skipping connections, or tuning certain hyper-parameters have limited effect over its performance. As such, many practitioners spend countless hours creating and curating annotated data to train the latest network architectures at the penalty of algorithm development. In addition, dataset creation is one of the most costly and demanding components of the entire computational pipeline~\cite{cit16,cit17,cit18}. Therefore, having good practices in place for data annotation is critical to ensure successful outcomes.\\
Even though computer vision scientists and engineers are continuously researching novel methods to reduce the dependency of computational models on annotated data (e.g. deep reinforcement learning, active learning, semi/self/un-supervised learning, adversarial learning techniques, etc.); when it comes to computer vision deployment and commercialization, supervised techniques continue to be the go-to approach. Massive data labeling efforts are being sponsored in commercial, government and academic environments. Open source annotated datasets are among the most desired assets sought by computer vision, machine learning, and deep learning scientists. As a consequence of being in large demand, when an new dataset is released, it will often quickly become a gold standard for a given application or use case. Examples of popular computer vision datasets include ImageNet~\cite{imagenet}, MS-COCO~\cite{mscoco}, VisualGenome~\cite{visualgenome}, SpaceNet~\cite{spacenet}, IARPA's Multi-view Stereo~\cite{mvs}. Datasets are equally used for both training the networks and measuring performance by running k-fold cross validation experiments on them. Therefore, the effects of data quality have implications in both the actual model performance and its evaluation \cite{Lampert}.\\
Data annotation is a manual process. In some instances, there are some automated bootstrapping routines to estimate or cluster data into labels, but in general, annotation is largely manual where tens, hundreds, or thousands of annotators label images according to a set of instructions. Annotation is done in a way such that a neural network can then learn a visual task (e.g. image segmentation, object detection, image classification, image captioning, etc.) by training on this data.\\
Even though modern neural network architectures are robust to noisy labels, consistent lack of quality should be a cause for concern. Careful thought should be given to blindly trusting open source data to be used for algorithm training and/or performance evaluation. There are many sources of error that can impact the quality of labeled data including lack of proper data management, instruction ambiguity, data misinterpretation due to low signal-to-noise ratio in the source data, lack of annotator focus, just to name a few. There is a need for instrumenting these frameworks with proper mechanisms to monitor data quality as labeling efforts progress.\\
With this in mind, in this work, we set to study the quality of annotated data and its impact to algorithm performance in computer vision so that systems trained with such data can be trusted. We acknowledge that this is a broad topic that can be extended in many research directions. As such, we have constrained it to the problem of object detection and focused on answering some questions that an organization may face during the planning stages of data annotation exercises.\\
Computer vision has become a success story within artificial intelligence, and in many cases is at the forefront of innovation. With object recognition and detection being an important part of it. Therefore, we believe that the object detection problem can be an excellent proxy for the broader challenge of data quality within AI.\\
An object detector is an algorithm that has been trained to locate known objects in an image, highlight the object's location with a bounding box, and label the object based on a relevant semantic category (e.g. dog, chair, vehicle, etc.). We have conducted a series of bounding box annotation experiments to quantify data quality in object detection image annotation and the effects of noisy labeled data to a state-of-the-art detector, \textit{RetinaNet} \cite{retinanet}. In particular, we are interested in shedding light on the following topics: data quality impacts from having a professional team of annotators vs. crowd sourced efforts, annotation agreement among annotators, metrics that can be used to quantify labeling quality, and data quality impacts on model performance. Our ultimate goal is to assess how these factors play a role in AI trustworthiness during human-machine interactions.\\
One important outcome from this work is that, in addition to knowledge of the training methodology being used, an understanding of the data labelling process and knowledge of ground truth data used for model evaluation are also vitally important; and each plays a major role in the trustworthiness assessment of an AI system. In addition, oversight in any of these components can have both performance and trust implications.\\
We can summarize our contributions as follows:
\begin{itemize}
    \item We have extended Krippendorff's Alpha for measuring annotator agreement for computer vision labeling.
    \item We have adapted a series of data quality control metrics based on Krippendorff's Alpha for object detection.
    \item We conducted a series of experiments that showed the challenges of treating any and all available annotated data as a black box without incorporating quality measures.
    \item We showed how inter annotator agreement can be used to select training data that leads to models with higher precision for the task of object detection.
\end{itemize}
The paper summarizes our experiments and findings as follows: section 2 reviews related work, sections 3 and 4 describe the annotation experiment and quality evaluation metrics. In section 5 we include the main results obtained in the experiments. Finally, we discuss some observations, and conclude the paper in sections 6 and 7.\\

\section{Related Work}
Due to its importance, data quality and annotation strategies have been studied in recent years in other areas of artificial intelligence. In McCulloh (2018), authors presented several metrics to measure and monitor performance and quality in large annotation teams for the task of natural language processing. The authors introduced measures for quantifying inter annotator agreement, rater consistency, distraction rate, among others \cite{McCulloh}.\\
In the area of image processing and computer vision, others have studied effects of crowdsource annotation. In Chen (2016), the authors conducted a study to assess the task completion time and accuracy of employing non-expert workers to process large datasets of images. They concluded that the crowd is more scalable and more economical than other solutions \cite{Chen}. In Castro (2018), the authors showed that for relatively unambiguous visual tasks one trained worker annotating each image and/or video is enough for obtaining high quality annotations as opposed to several workers \cite{Castro}. In Walker (2018), Walter and Sorgel highlighted common problems in data annotation for geospatial applications including object misrepresentations. They showed that by annotating data multiple times and using one common data set some of these quality inconsistencies disappear \cite{Walker}.\\
Finally, others have studied the effect of annotated data quality on actual algorithm performance measurements. Besides algorithm training, annotated datasets are also used as ground truth data for algorithm benchmark platforms. In Lampert (2016), the authors showed that for image segmentation several algorithms can be ranked differently depending on how the ground truth dataset is collected. In a somewhat concerning statement, authors concluded that in some ground truth datasets it is not possible to confidently benchmark an algorithm \cite{Lampert}. 

\section{Image Data Annotation}
Two publicly available data-sets were used for the image annotation task. Both datasets are a collection of overhead videos that share some of the same object categories: 

\begin{itemize}
    \item \textbf{VIRAT Video Dataset}. These videos consist mainly of stationary ground-based camera data of outdoor spaces such as parking lots, streets, and construction sites \cite{VIRAT}. The VIRAT videos were annotated on a frame-by-frame basis.    
    \item \textbf{Stanford Drone Dataset} from the Computational Vision and Geometry lab at Stanford University. The drone videos were split by our team into images that were used for annotation.
     \item \textbf{Semantic Drone Dataset} (SDD-ICGV) from the Institute of Computer Graphics and Vision in the Graz University of Technology. These images focus on urban scenes from a nadir perspective. Their training set has bounding box annotation in over 20 classes including our three categories of interest \cite{SDD-ICGV}.
    \item \textbf{Berkeley Deep Drive 100K} (BDD100K) from the Berkeley Deep Drive Industry Consortium. These images originated 100,000 HD video sequences of driving experience that was split into images and annotated with 2D bounding boxes \cite{BDD100K}.
\end{itemize}

\begin{figure}[ht] 
  \begin{subfigure}[b]{0.5\linewidth}
  \centering
    \includegraphics[width=.9\linewidth]{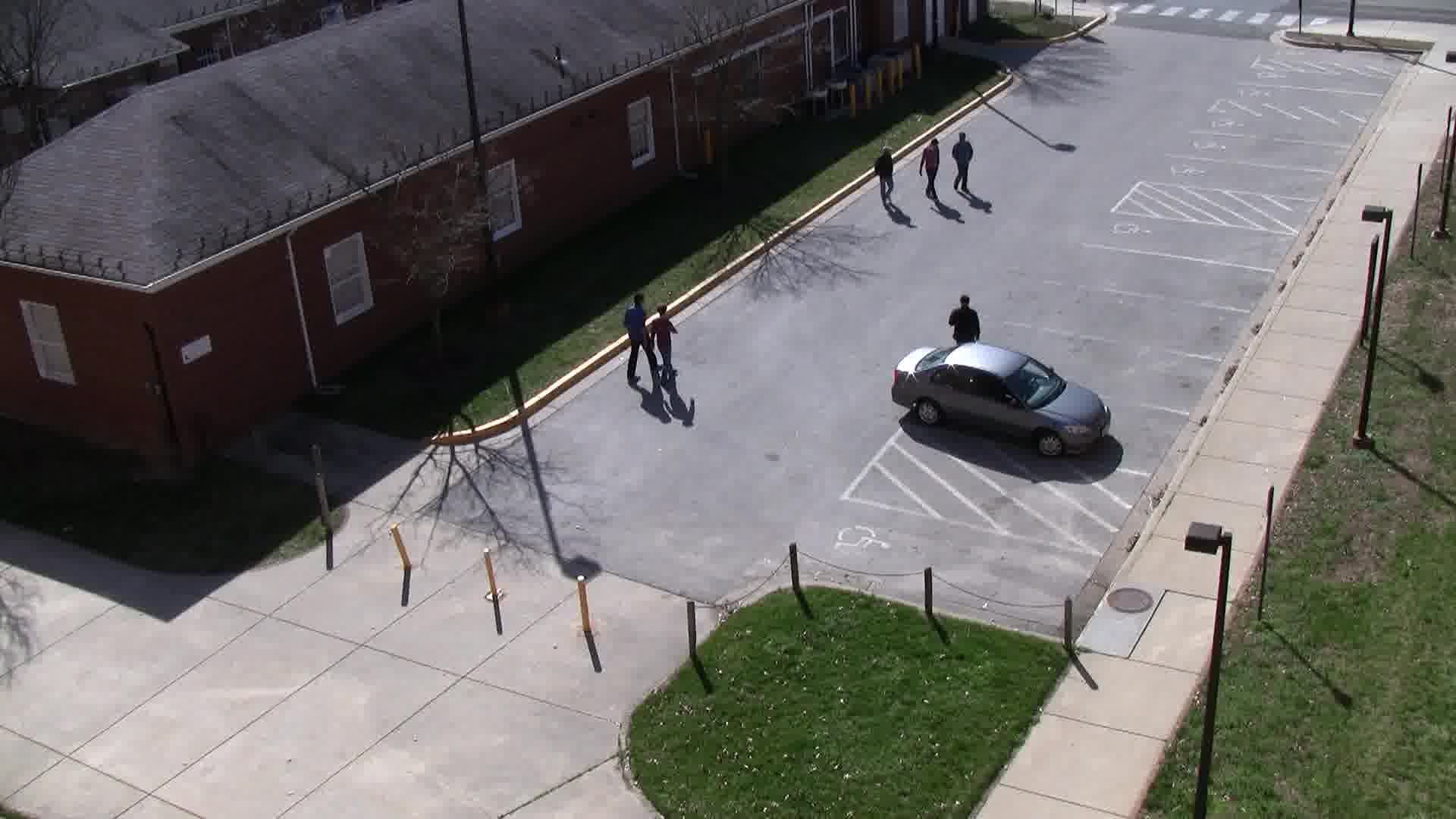} 
    \caption{VIRAT} 
	\label{fig:a} 
	\vspace{4ex}
  \end{subfigure}
  \begin{subfigure}[b]{0.5\linewidth}
  \centering
    \includegraphics[width=.9\linewidth]{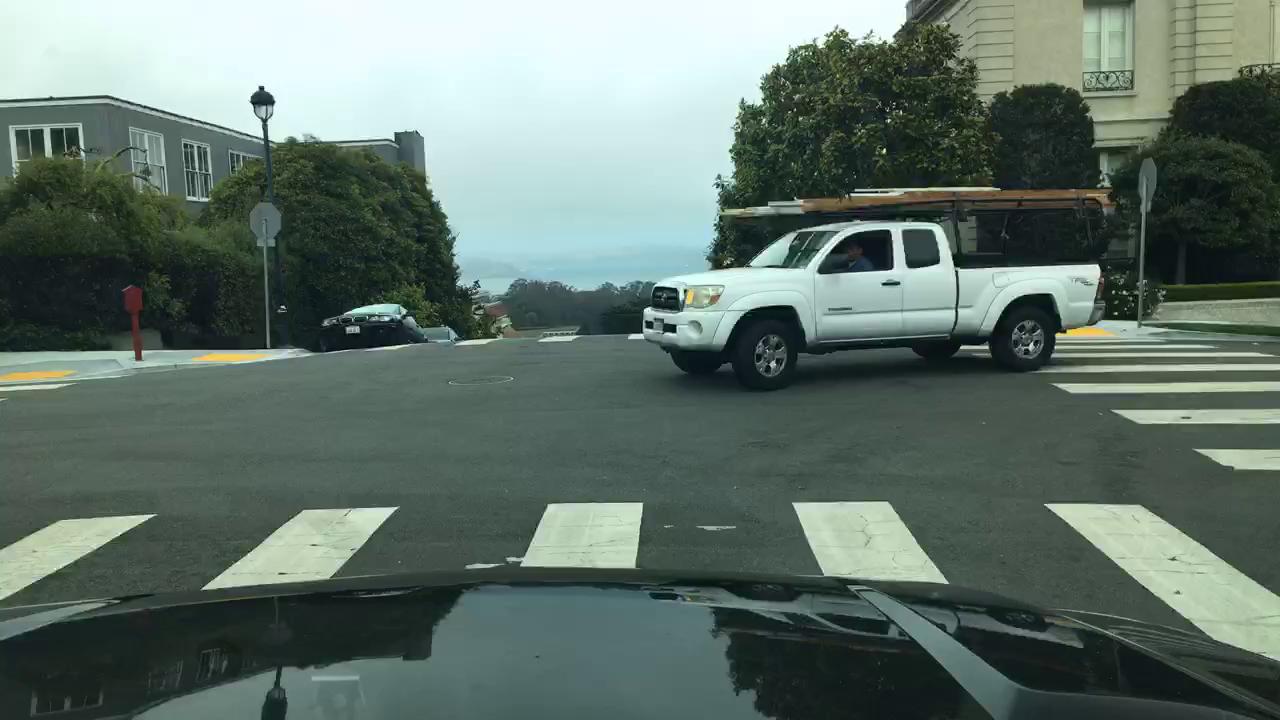} 
    \caption{BDD}
	\label{fig:b}
	\vspace{4ex}
  \end{subfigure} 
  \begin{subfigure}[b]{0.5\linewidth}
  \centering
    \includegraphics[width=.9\linewidth]{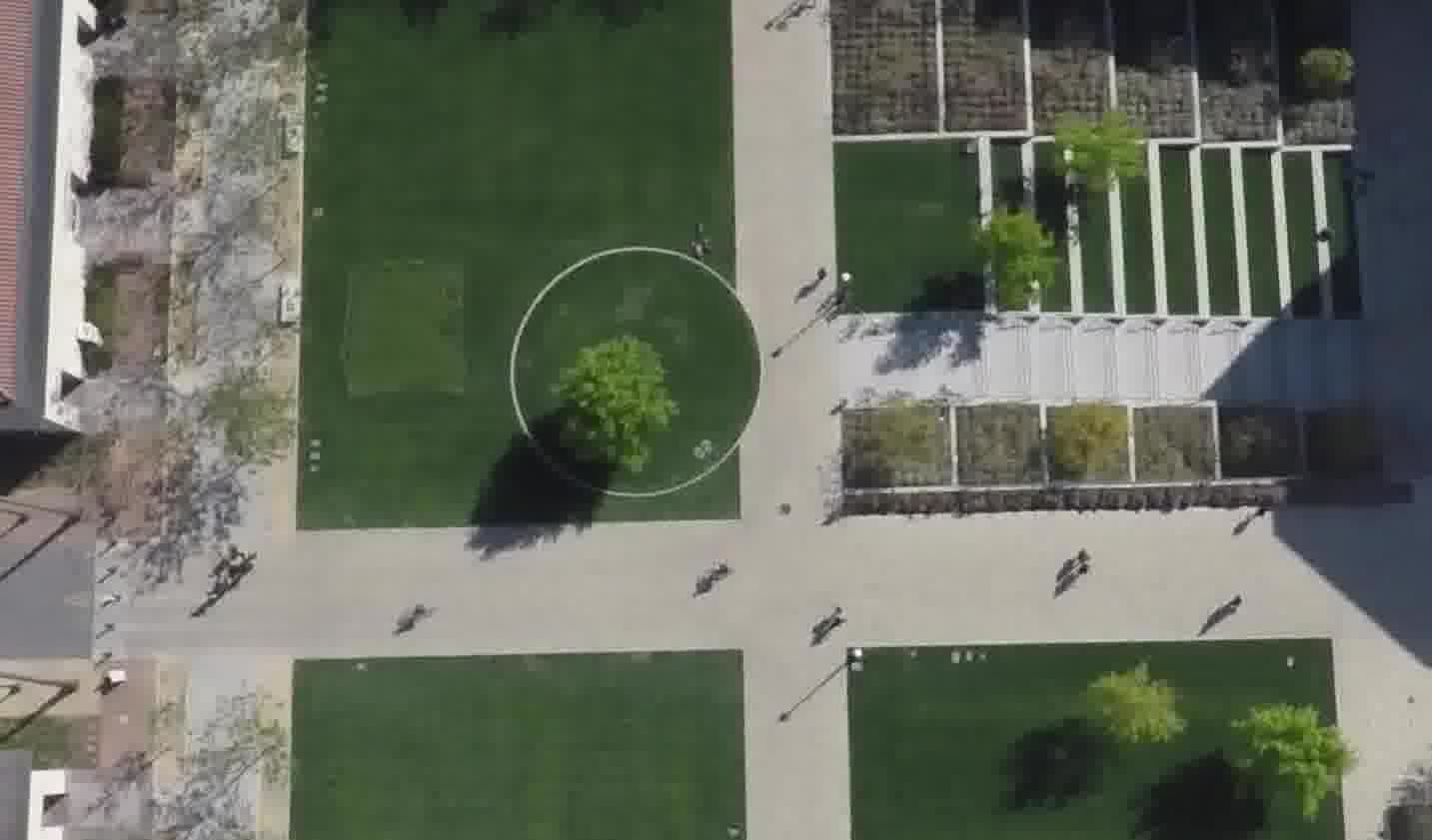} 
    \caption{SDD} 
	\label{fig:c}
	\vspace{4ex}
  \end{subfigure}
  \begin{subfigure}[b]{0.5\linewidth}
  \centering
    \includegraphics[width=.9\linewidth]{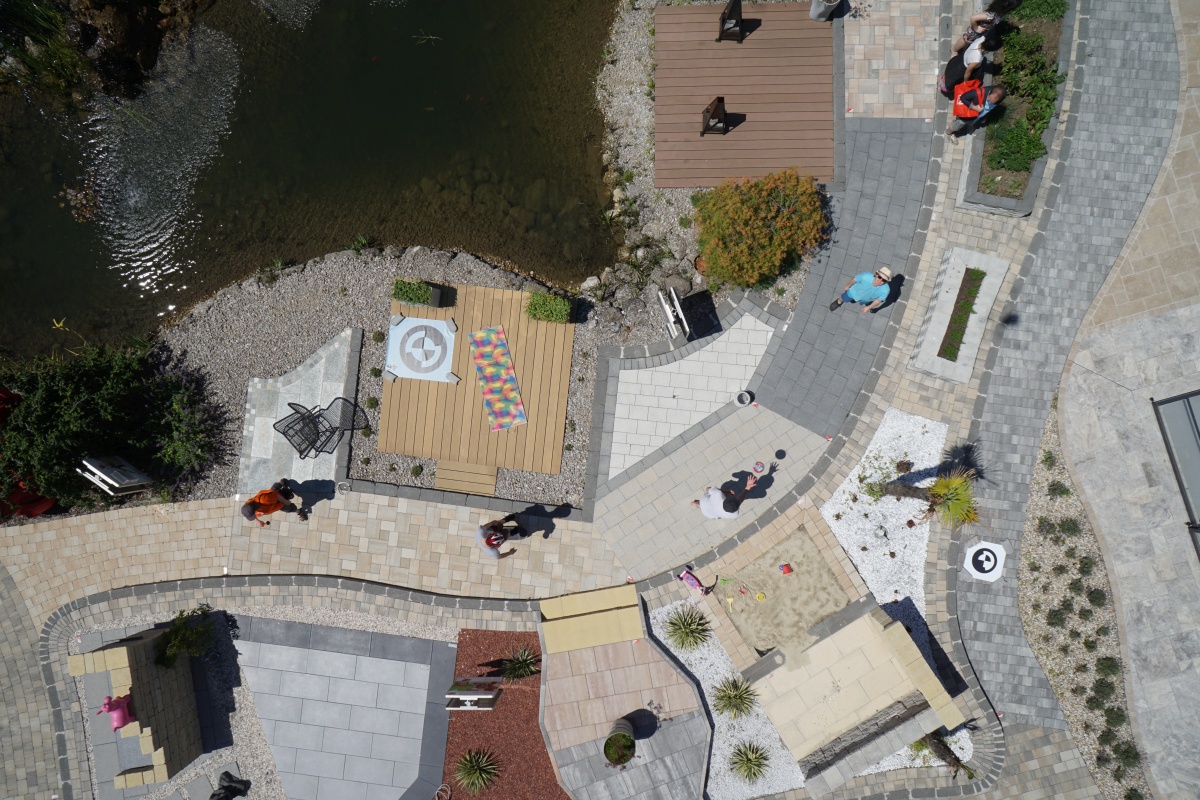} 
    \caption{SDD-ICGV}
	\label{fig:d}
	\vspace{4ex}
  \end{subfigure}
\caption{Dataset Samples}
\label{fig}
\end{figure}
The images selected for annotation were taken from high and low altitudes and from ground-level in order to incorporate variability in bounding box size. The quality metrics described below were calculated on each image separately to account for differences in image scale. The low altitude images were recorded at an altitude of 5-30 meters, while the high altitude images were recorded at 30+ meters. Bounding boxes ranged from 25x25 pixels to 1278x220 pixels. The differences in altitude and resolution contributed to the wide range of boxing box sizes. Additionally, images from the SDD-ICGV data source were re-scaled to 1200 x 800 pixels before they were annotated.\\
During the annotation task, annotators were asked to identify people, vehicles, and bicycles in a given image or frame. If found, they would draw a bounding box around the specific object and select a corresponding category label. The tasks consisted of finding objects of interest, then labeling the objects with their appropriate category type. The difference in images made this a more complicated task given that the resolution of some of the images made it difficult to distinguish between people and bicycles in some cases. Other factors of disagreement where cluttered scenes where some objects were missed by some annotators, the spatial extent of the box for certain objects, some annotators draw very tight boxes, while others included additional background pixels due to non rigidity of some of the targets.\\
\begin{figure*}[tb]
  \centering
  \includegraphics[width=\textwidth]{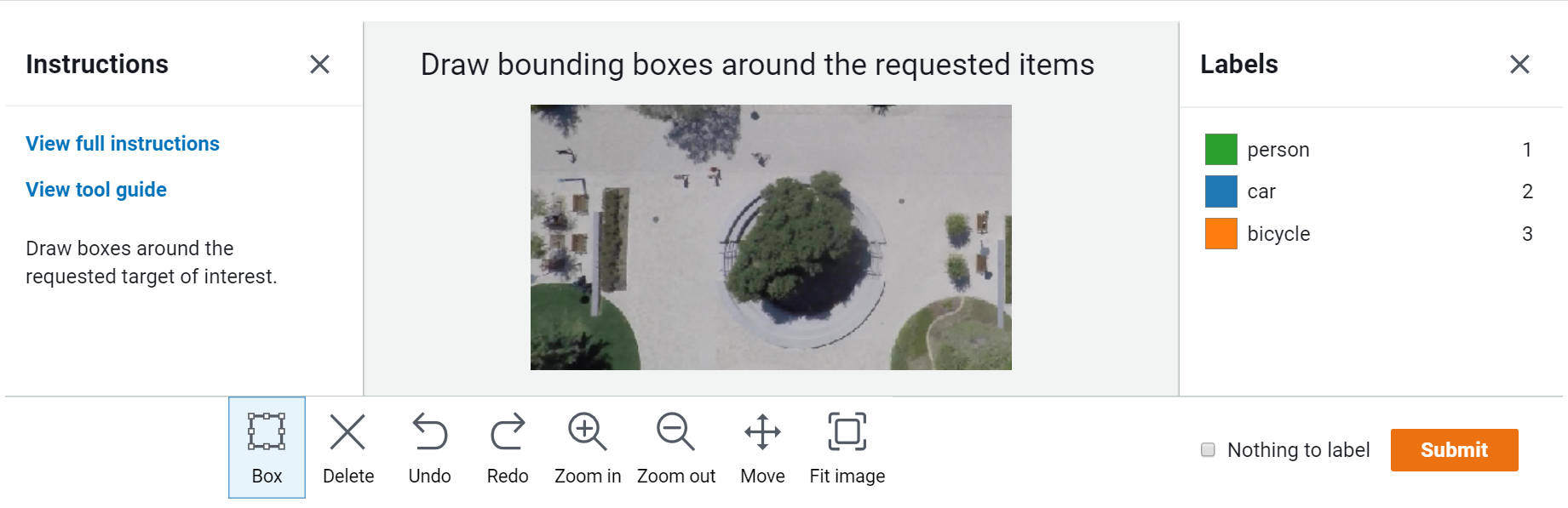}
  \captionof{figure}{Amazon Mechanical Turk Annotation Layout.}
\end{figure*}

The instructions given included: draw the bounding box tight around each object, minimizing empty space; draw boxes that cover each object completely even if this box overlaps other objects; do not include parts of the object that cannot be seen; avoid including shadows; if the target is off the screen, draw the box up to the edge of the image; and, include as many boxes of object as you can see in the image.\\
We considered the two most popular types of workforce for annotation: a team of professional annotators, and crowdsource resources. 

\begin{enumerate}
    \item Professional annotators. The first group consisted of four trained professional annotators with Geospatial Information System experience. This group has experience viewing satellite and low resolution images as well as identifying objects from these images. They had direct supervision and had daily meetings reporting their progress and any issues that may have come up. This group had the opportunity to interact directly with each other and their supervisors, but they were asked not to speak to each other about their tasks or the images being annotated. This group annotated batches of images using their own platform, and sent results back to our team as they were completed.  
    \item Crowdsourced annotators. The second group of annotators came from Amazon Mechanical Turks (MTurk), an online marketplace for outsourcing work \cite{Buhrmester2011}. Given the diversity of workers (turkers) available, these coders were divided into three categories: novice, experienced, and expert according to the criteria in Table \ref{MTurkQ}.  
\end{enumerate}

\begin{center}
\begin{tabular}{p{1.6cm}p{1.5cm}p{1.5cm}p{1.5cm}}
 \hline
 \multicolumn{4}{|c|}{MTurk Annotator Criteria} \\
 \hline
 Category & Number of Tasks (HITs) Completed & Approval Min & Approval Max\\
 \hline
 Novice   & 50    & 75\% &   100\%\\
 Experienced &   500  & 85\%   & 99\%\\
 Expert & & 99\% &  100\%\\
 \hline
\end{tabular}
  \captionof{table}{MTurk Qualification Requirements}
  \label{MTurkQ}
\end{center}    

The turkers were given basic instructions on how to draw bounding boxes and how to choose a  corresponding category label. 

In order to capture inter-coder agreement and cross group comparison metrics, all groups were given the same set of images to annotate. Each image that was uploaded to MTurk was annotated ten times by each of the 3 MTurk groups. The same images were also annotated once by each coder from the professional annotation team. This means that a given image was annotated by four professional, and thirty times by turkers (10 times from each of the three different mturk groups.  

\begin{figure}[ht] 
  \begin{subfigure}[b]{0.5\linewidth}
  \centering
    \includegraphics[width=.9\linewidth]{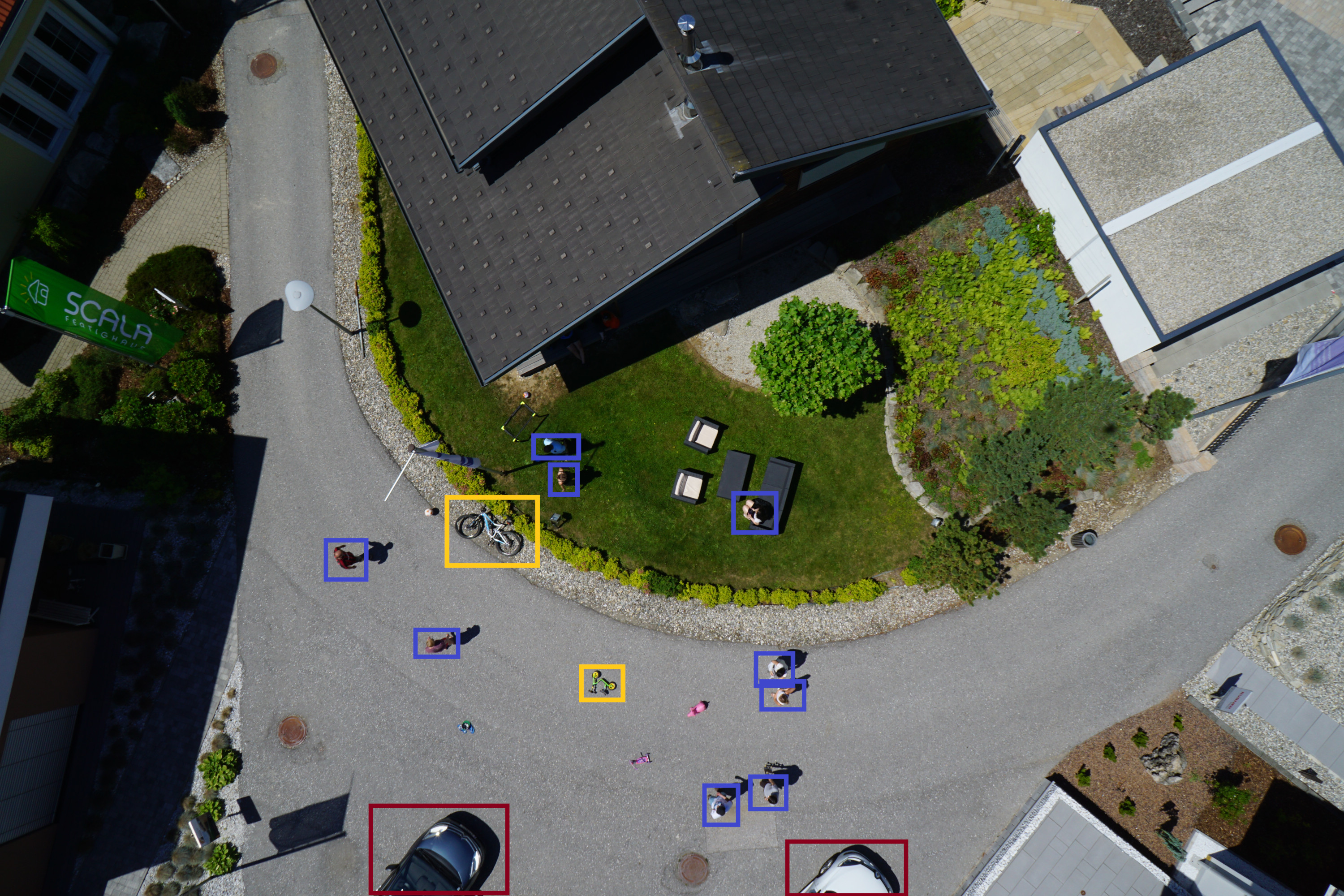} 
    \caption{Good Annotation} 
	\label{fig2:a} 
	\vspace{4ex}
  \end{subfigure}
  \begin{subfigure}[b]{0.5\linewidth}
  \centering
    \includegraphics[width=.9\linewidth]{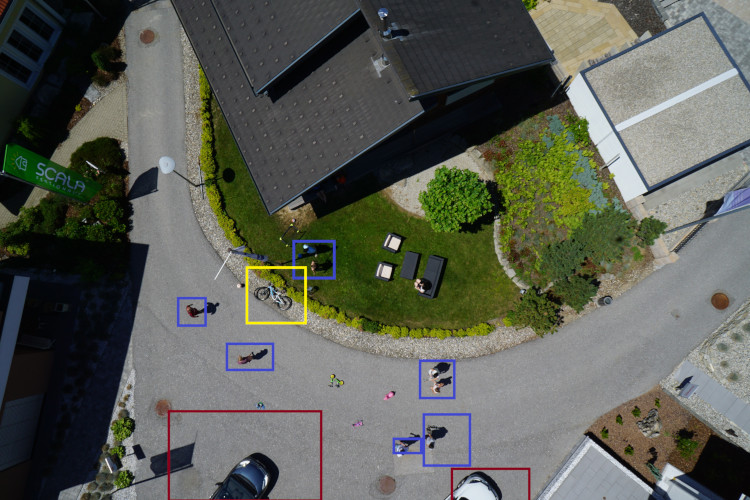} 
    \caption{Bad Annotation}
	\label{fig2:b}
	\vspace{4ex}
  \end{subfigure} 
  \begin{subfigure}[b]{0.5\linewidth}
  \centering
    \includegraphics[width=.9\linewidth]{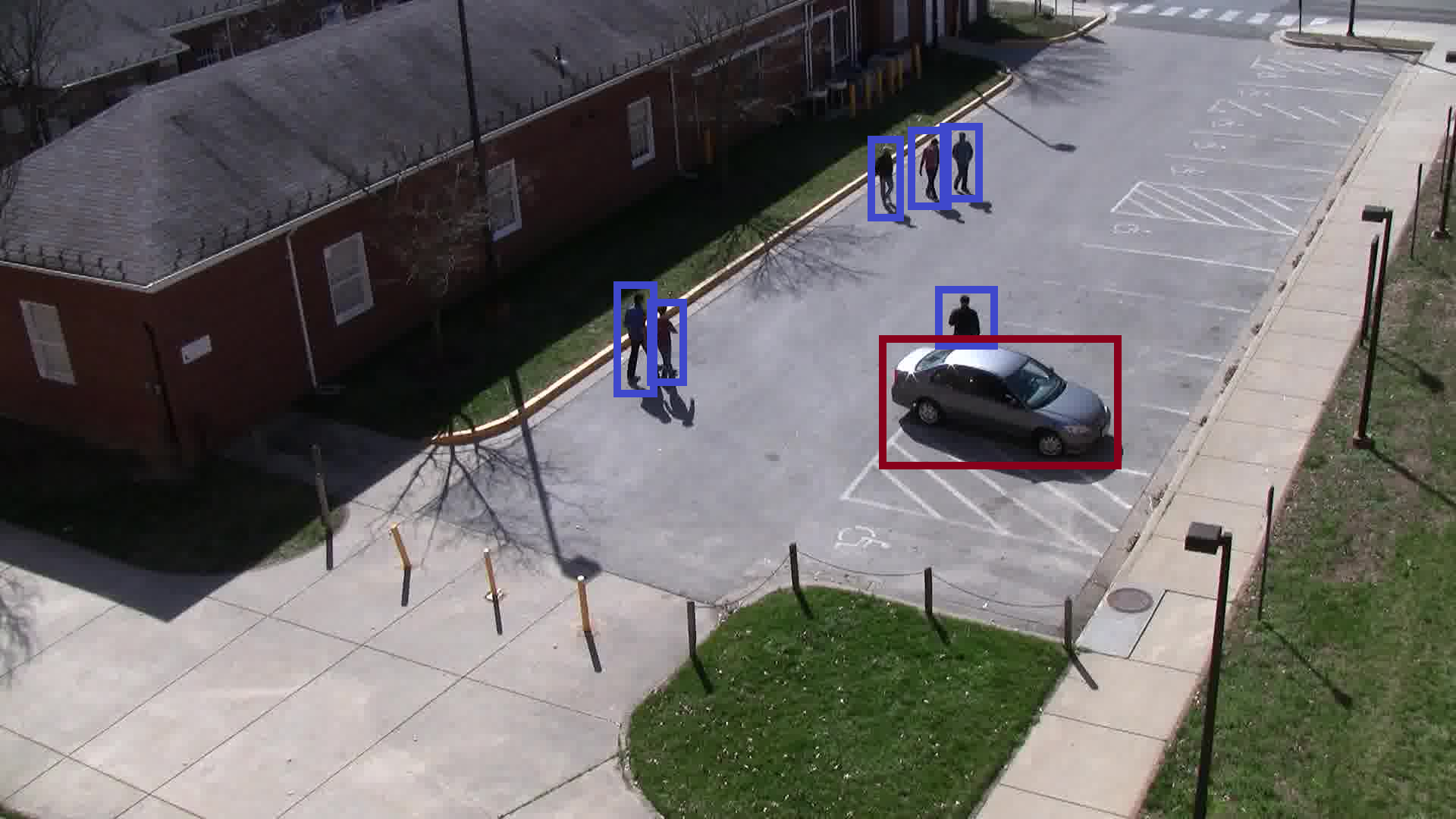} 
    \caption{Good Annotation} 
	\label{fig2:c}
	\vspace{4ex}
  \end{subfigure}
  \begin{subfigure}[b]{0.5\linewidth}
  \centering
    \includegraphics[width=.9\linewidth]{bad2s.png} 
    \caption{Bad Annotation}
	\label{fig2:d}
	\vspace{4ex}
  \end{subfigure}
    \caption{Good \& Bad Annotation Examples}
\label{GoodBad} 
\end{figure}

\section{Data Quality and Model Performance Evaluation Metrics}
In this section we describe the metrics used to measure the quality of annotations from the two different annotator groups. When developing annotated datasets, it is important to track quality as labeling progresses so that corrective measures can be introduced. We distinguish between two types of metrics: Data quality metrics and algorithm performance metrics.
\subsection{Data Quality Metrics}
Data quality metrics aim at measuring the accuracy of the labels. In the case of object detection, this refers to measuring the correctness of the semantic label (e.g. person, vehicle, bicycle) and the accuracy of each object's bounding box placement in the image. Working under the assumption that annotators have a shared understanding of what and how data should be annotated, we believe that collective knowledge should provide the most accurate reference framework to evaluate labels. Therefore, we measure (relative) accuracy for each annotator against the group. The relative accuracy produced by a group of individual annotators is measured by computing Inter-Annotator Agreement (IAA) metrics. In this case, we proposed to use Krippendorff's Alpha ($K\mbox{-}\alpha$) \cite{krip1,krip2} as it handles multiple raters, missed bounding boxes, and box size variation.
We have extended $K\mbox{-}\alpha$ for computer vision annotations according to:

\begin{equation}
\alpha = 1 - \frac{D_o}{D_e} = 1-\frac{\sum_{x=1,x'=1}^X o_{xx}\delta(x,x')}{\frac{1}{n-1} \sum_{x=1,x'=1}^Xn_x n_{x'}\delta(x,x')}
\end{equation}

Where $D_o$ is the disagreement observed, and $D_e$ is the disagreement expected by chance. $X$ represents the categories of labels (e.g. person, vehicle, bicycle). $x$ and $x^{'}$ are the labeling options of two different annotators. $o$ represents the pair of labels that were given by two different annotators, $n_{x} \text{ and } n_{x^{'}}$ are the number of annotations observed for each class, with $n$ being the sum of pairwise values among annotators. $\delta{(x,x^{'})}$ is the difference function between $x$ and $x^{'}$, where

\[
    \delta{(x,x^{'})} = 
\begin{cases}
    0 \text{ if } x = x^{'}\\
    1, \text{ if } x \neq x^{'} 
\end{cases}
\]

The random choice benchmark in $K\mbox{-}\alpha$ is a function of the number of annotators. As the number of annotators gets very large, the random choice benchmark converges to $0$ \cite{McCulloh}. The range of $K\mbox{-}\alpha$ is $-1$ to $1$. An industry and academic benchmark for good IAA using $K\mbox{-}\alpha$ is $0.67$, with $0.8$ and above being very good \cite{krip1}.\\
In the case of the annotated images, each pixel was taken as an individual observation and calculated in two ways. First pixels were assigned a binary value within a multi-dimensional (3D) array; in each 2-D sub-array, a pixel was either labeled (residing inside a bounding box), or unlabeled (residing outside of any bounding box). The information for each class was encoded in the third dimension or channel -each class was assigned to a separate channel and every pixel was mapped to either a labeled or unlabeled value within a given channel according to the criteria used above. The 3-D pixel label arrays were flatten into a one dimensional array, and $\frac{1}{5}$ of the pixels were randomly dropped to reduce the computational time.\\
In addition to the $K\mbox{-}\alpha$ extension, we adapted data quality metrics presented in \cite{McCulloh} to the annotation of images for object detection. The metrics used to assess annotation quality are \textit{Rater Vitality} and \textit{Recognition Difficulty}.
\begin{enumerate}
\item
\textit{Rater Vitality} ($V_{i}$), which show how well each annotator agrees with the larger group of annotators, and the impact that individual's annotations have on overall consensus. Rater Vitality is calculated by the following:
\[ V_{i} = K_{\alpha} - K_{i}\]
Where $V_{i}$ is the vitality rating for annotator $i$. $K_{\alpha}$ is the overall Krippendorff's Alpha including all annotators for a given observation, and $K_{i}$ is Krippendorff's Alpha for all annotators except annotator $i$. If an annotator’s vitality score is positive, then their annotations positively impacted overall consensus, and if an annotator's vitality score is negative, then they decreased overall consensus.

\item \textit{Class Recognition Difficulty}
We define Class Recognition Difficulty, $V_i(\lambda)$, as the \textit{Rater Vitality Score} calculated from the IAA within individual classes ($\lambda$). Recognition Difficulty can be used for comparison to overall IAA. A greater degree of disagreement within a class can suggest that a particular class is more difficult to annotate, or annotators need further guidance on identifying the class. In the later case, $V_{i}$ of a given annotator $i$ can be helpful in identifying where an individual annotator or sub-group has disagreement with the majority. 
\end{enumerate}


\subsection{Model Performance Metrics}
In addition to annotation quality evaluation we wanted to investigate its connection with algorithm performance. For object detection, well accepted metrics include precision, recall and F1 score. F1 is the harmonic mean between precision and recall. It tends to capture the lower value of the two. If either recall or precision are small, the other one does not matter too much since F1 will reflect the low value. On the contrary, if F1 has a high score, both precision and recall indicate good results. More formally, F1 is computed as:
\begin{equation}
F1-score = \frac{2 \cdot precision\cdot recall}{precision+ recall}
\end{equation}
with,
\begin{equation}
precision=\frac{TP}{TP+FP};recall=\frac{TP}{TP+FN}
\end{equation}
$TP$ refers to true positives, $FN$ and $FP$ are false negatives and false positives respectively. What remains is to define a true positive when comparing bounding boxes. We used Intersection over Union (IoU) to obtain the degree of overlap between two boxes (annotator and ground truth). IoU allows us to evaluate how similar a predicted box is to a ground truth bounding box by comparing the ratio of the area where two boxes, with the same semantic label, overlap to the total combined area of the two boxes. In our case, an area is described by the number of pixels in the region of interest. The larger the ratio, the better is the match between ground truth and the annotator box. In case the label assigned to the box is different from the ground truth, IoU is returned as zero regardless of the level of overlap between the two boxes. Only when the IoU value exceeds a predetermined threshold, it is considered a true positive.\\
\textbf{Selection of the ground truth dataset}: one important aspect is the selection of the appropriate ground truth reference to measure the algorithm performance. When a new dataset is open sourced, many times annotated data is included as part of the release. It often comes with annotation errors as reported by others \cite{imagenet}, \cite{gterror1}, \cite{gterror3}, \cite{Lampert}. In many situations, these datasets are annotated for training purposes rather than for evaluation. Noisy labels are less critical during training than for rigorous algorithm performance since neural networks can tolerate and ``welcome'' noise and data diversity. As a consequence, some errors in labeling go unnoticed during performance evaluation.\\
We have compared the algorithm performance using the labels included in the dataset original release and custom datasets derived from the findings in the inter annotator agreement and vitality experiments. To do so, we studied the level of agreement among annotators and removed data from annotators with a lower level of agreement with the group from the evaluation dataset.\\ 
\section{Experimental Results}
In this section we described our experiments for both data annotation quality monitoring and algorithm performance.
\subsection{Data Annotation Quality} 
Using the procedure described in section 4, $K\mbox{-}\alpha$ was calculated separately for teams of annotators and data sets annotated. For the professional team, IAA and quality metrics were calculated from the annotations conducted by all 4 workers. 93 images were used from the Stanford Drone data set, and 125 images from the VIRAT data set. The average and median $K\mbox{-}\alpha$ values are found in Table \ref{ratings}.\\

\begin{center}
\begin{tabular}{ccc}
 \hline
Data Set & Average $K\mbox{-}\alpha$ & median $K\mbox{-}\alpha$ \\
 \hline
Stanford Drone & 0.575 &  0.582 \\
 \hline
 VIRAT & 0.881 & 0.889 \\
 \hline
\end{tabular}
  \captionof{table}{Average and median $K\mbox{-}\alpha$ rating for images annotated by the professional team.}
  \label{ratings}
\end{center}

The quality metrics calculated for each of the 4 annotators in the professional team can be seen in Table \ref{vitality} and Table \ref{recognition}.\\

\begin{center}
\textbf{Vitality Score} \\
\begin{tabular}{p{1.6cm}p{1cm}p{1cm}p{1cm}p{1cm}}
 \hline
& Drop Antr 1 & Drop Antr 2 & Drop Antr 3 & Drop Antr 4 \\
 \hline
 \multicolumn{5}{c}{Stanford Drone} \\
 \hline
Average $V_{i}$ & -0.078 & 0.028 & 0.012 & 0.022 \\
Median $V_{i}$ & -0.069 & 0.026 & 0.010 & 0.018 \\
\hline
 \multicolumn{5}{c}{VIRAT} \\
\hline
Average $V_{i}$ & -0.002 & -0.001 & 0.007 & -0.003 \\
Median $V_{i}$ & 0 & 0 & 0.006 & 0 \\
\hline
\hline
\end{tabular}
  \captionof{table}{Average Vitality score of annotators on the professional team.}
  \label{vitality}
\end{center}

\begin{center}
\textbf{Class Difficulty} \\
\begin{tabular}{p{1.1cm}p{1.1cm}p{1cm}p{1cm}p{1cm}p{1cm}}
 \hline
Class & Avg. $K\mbox{-}\alpha$ & Drop Antr 1 & Drop Antr 2 & Drop Antr 3 & Drop Antr 4 \\
 \hline
 \multicolumn{6}{c}{Stanford Drone} \\
 \hline
Person & 0.463&-0.124&0.029&-0.004&0.050 \\
Vehicle &0.675&0.010&0.048&-0.095&0.049 \\
Bicycle & 0.408 &-0.045&0.056&0.015&-0.023\\
\hline
 \multicolumn{6}{c}{VIRAT} \\
\hline
Person & 0.817&-0.010&0.007&0.003&0.009 \\
Vehicle & 0.897 &-0.004&-0.004&0.011&-0.006\\
Bicycle & 0.197&0.036&-0.051&0.012&0.145\\
\hline
\hline
\end{tabular}
  \captionof{table}{Average Vitality score of each annotator calculated on individual classes}
  \label{recognition}
\end{center}
To baseline the professional team with crowd sourced annotators, a few of the images annotated by the professional team were selected from the Stanford Drone data set and given to MTurk workers to annotate. Each image was annotated by 10 workers from each MTurk group (expert, experienced, novice). $K\mbox{-}\alpha$ was calculated from all 10 workers in each experience group. The average and median $K\mbox{-}\alpha$ values for each group are found in Table \ref{mbox}.\\
\begin{center}
\begin{tabular}{cccc}
 \hline
$K\mbox{-}\alpha$ & Expert & Experienced & Novice\\
 \hline
Average & 0.310 & 0.371 &  0.086 \\
Median & 0.269& 0.392 & 0.059 \\
 \hline
\end{tabular}
  \captionof{table}{Average and median $K\mbox{-}\alpha$ rating of MTurk workers for Stanford Drone images annotated by both the professional team and MTurk workers}
  \label{mbox}
\end{center}
\subsection{Algorithm Performance Evaluation}        
We conducted a series of experiments with the goal of understanding the effect of training data selection on algorithm performance given the available annotated data. For all the algorithm performance experiments we used the data annotated by the team of professionals (4 annotators). Stanford Drone Dataset was used. We trained RetinaNet with four different datasets by leaving one annotator out each time as we did with the data quality experiments (see Table \ref{vitality}). Thus, for dataset \textit{Drop Antr $k$} we left out the k-th annotator group. In our case we studied the individual contribution and the level of agreement per annotator so each annotator group consisted of one single annotator. For example, dataset 1 (\textit{Drop Antr 1}) consisted of data labeled by annotators 2, 3, and 4. Without loss of generality, this can be extended to larger groups and subgroups of annotators that can be segmented in different manners according to IAA analysis.\\
\paragraph{Training Details} As mentioned earlier, we compared the performance of RetinaNet algorithm \cite{retinanet} for object detection. We focused on the annotations produced by trained annotators on the Stanford Drone Dataset. We trained the detector with datasets \textit{Drop Antr 1}, \textit{Drop Antr 2}, \textit{Drop Antr 3}, and \textit{Drop Antr 4}, described above, to match the data quality analysis. The goal is to observe the effect of measuring data quality and how it translates to algorithm performance.\\
RetinaNet was trained for 10 epochs, using more than 11,000 sample instances of vehicles, people and bicycles. We used ResNet-50 as the backbone, and fine-tuned the entire network on top of a pre-trained version with MS-COCO \cite{mscoco}. We turned off data augmentation since we wanted to focus only on the effect of the manually generated annotations. The test set included more than 1000 object instances.\\
\paragraph{Ground truth selection} For any algorithm evaluation, there is the question of validation or ground truth data selection. One approach is to use data annotated from higher performing annotators according to IAA and vitality results (Table \ref{vitality}). In the case of the professional team, annotators 2, 3, and 4 showed higher degree of agreement. Therefore, we randomly selected test data annotated from these annotators and gathered two custom ground truth datasets. One ground truth dataset was created from a random selection among all three annotators (\textit{gt1} dataset), and the other was used from data of the same annotator selected at random from the top 3 \textit{gt2}. In addition to these two ground truth datasets, we added the original labels released with the Stanford Drone data \textit{gt3} to the comparisons. The goal was to see the effects of using data that has been validated through data quality analysis vs. another set that is treated as a black box and had no visibility on its quality control. In the first experiment we compared the algorithm (RetinaNet) performance against \textit{gt3} dataset. Table~\ref{tab:resultsgt} shows RetinaNet performance for this set.\\
\begin{center}
\begin{tabular}{cccc}
 \hline
 Training Dataset for RetinaNet & Prec. & Rec. & F1\\
 \hline
 Drop Antr 1 & 0.41  & 0.49 & 0.42 \\
 \hline
 Drop Antr 2 & 0.38 & 0.61   & 0.46\\
 \hline
Drop Antr 3 & 0.40 & 0.62 &  0.49 \\
 \hline
 Drop Antr 4 & 0.41 & 0.63 &  0.50\\
 \hline
\end{tabular}
  \captionof{table}{RetinaNet trained with 4 datasets (\textit{Drop Antr 1}, \textit{Drop Antr 2}, \textit{Drop Antr 3}, and \textit{Drop Antr 4}) performance measured against original ground truth (\textit{gt3} dataset).}
  \label{tab:resultsgt}
\end{center}
Results from the same experiments but using ground truth sets \textit{gt1} and \textit{gt2} are shown in tables \ref{tab:results05gt1} and \ref{tab:results05gt2} respectively.

\begin{center}
\begin{tabular}{p{1.7cm}p{0.5cm}p{0.5cm}p{0.5cm}p{0.5cm}}
 \hline
Training Dataset for RetinaNet & Prec. & Rec. & F1 & mis-class.\\
 \hline
 Drop Antr 1 & 0.65    & 0.56 & 0.60 & 380\\
 \hline
 Drop Antr 2 & 0.60 & 0.67   & 0.63 & 781\\
 \hline
 Drop Antr 3 & 0.55 & 0.64 &  0.59 & 838\\
 \hline
 Drop Antr 4 & 0.58 & 0.66 &  0.62 & 738\\
 \hline
\end{tabular}
  \captionof{table}{RetinaNet trained with 4 datasets (\textit{Drop Antr 1}, \textit{Drop Antr 2}, \textit{Drop Antr 3}, and \textit{Drop Antr 4}) performance measured against custom ground truth \textit{gt1} dataset. Precision, recall, F1-score and number of mis-classified predictions for RetinaNet trained on four different sets of data.}
  \label{tab:results05gt1}
\end{center}

\begin{center}
\begin{tabular}{p{1.7cm}p{0.5cm}p{0.5cm}p{0.5cm}p{0.5cm}}
 \hline
Training Dataset for RetinaNet & Prec. & Rec. & F1 & mis-class.\\
 \hline
 Drop Antr 1 & 0.70 & 0.57 & 0.63 & 375\\
 \hline
 Drop Antr 2 & 0.60 & 0.66   & 0.63 & 847\\
 \hline
 Drop Antr 3 & 0.59 & 0.65 &  0.62 & 814\\
 \hline
 Drop Antr 4 & 0.63 & 0.67 &  0.65 & 711\\
 \hline
\end{tabular}
  \captionof{table}{RetinaNet trained with 4 datasets (\textit{Drop Antr 1}, \textit{Drop Antr 2}, \textit{Drop Antr 3}, and \textit{Drop Antr 4}) performance measured against custom ground truth \textit{gt2} dataset. Precision, recall, F1-score and number of mis-classified predictions for RetinaNet trained on four different sets of data.}
  \label{tab:results05gt2}
\end{center}
A somewhat expected observation was that RetinaNet models trained with data that included labels from annotator 1 (these are the sets labeled as \textit{Drop Antr 2}, \textit{Drop Antr 3}, and \textit{Drop Antr 4} in the experiments) produced a larger number of predictions - more than 40\% extra predictions. As a consequence, we ran another set of experiments, this time constraining the number of predictions (boxes) produced by each model to similar numbers. Tables \ref{tab:resultsconsgt1} and \ref{tab:resultsconsgt2} show performance results of RetinaNet using ground truth from the best annotators (i.e. \textit{gt1}, and \textit{gt2} sets).
\begin{center}
\begin{tabular}{p{1.7cm}p{0.5cm}p{0.5cm}p{0.5cm}p{0.5cm}}
 \hline
 Training Dataset for RetinaNet & Prec. & Rec. & F1 & mis-class.\\
 \hline
 Drop Antr 1  & 0.66  & 0.56 & 0.61 & 380\\
 \hline
 Drop Antr 2 & 0.62 & 0.56   & 0.59 & 456\\
 \hline
 Drop Antr 3 & 0.58 & 0.52 &  0.55 & 474\\
 \hline
 Drop Antr 4 & 0.61 & 0.54 &  0.57 & 422\\
 \hline
\end{tabular}
  \captionof{table}{RetinaNet trained with 4 datasets (\textit{Drop Antr 1}, \textit{Drop Antr 2}, \textit{Drop Antr 3}, and \textit{Drop Antr 4}) performance measured against custom ground truth \textit{gt1} dataset. Precision, recall, F1-score and number of mis-classified predictions.}
  \label{tab:resultsconsgt1}
\end{center}

\begin{center}
\begin{tabular}{p{1.7cm}p{0.5cm}p{0.5cm}p{0.5cm}p{0.5cm}}
 \hline
 Training Dataset for RetinaNet & Prec. & Rec. & F1 & mis-class.\\
 \hline
 Drop Antr 1  & 0.70 & 0.57 & 0.63 & 375\\
 \hline
 Drop Antr 2 & 0.60 & 0.59 & 0.59 & 625\\
 \hline
 Drop Antr 3 & 0.61 & 0.57 & 0.59 & 566\\
 \hline
 Drop Antr 4 & 0.64 & 0.59 & 0.61 & 504\\
 \hline
\end{tabular}
  \captionof{table}{RetinaNet trained with 4 datasets (\textit{Drop Antr 1}, \textit{Drop Antr 2}, \textit{Drop Antr 3}, and \textit{Drop Antr 4}) performance measured against custom ground truth \textit{gt2} dataset. Precision, recall, F1-score and number of mis-classified predictions.}
  \label{tab:resultsconsgt2}
\end{center}
\section{Discussion}
\begin{figure}[htb]
  \centering
  \includegraphics[width=0.4 \textwidth]{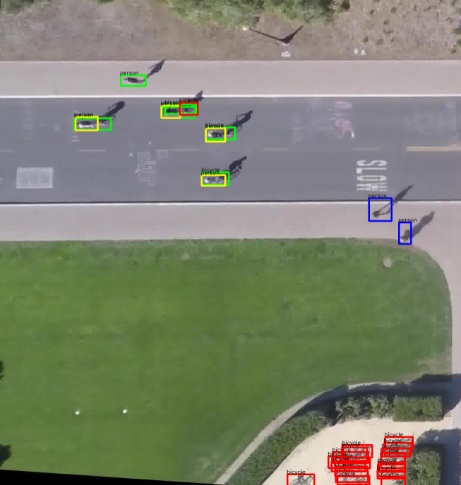}
  \captionof{figure}{Example of detection and ground truth mismatch. Red boxes indicate objects of interest not available in the ground truth that the model predicted, blue boxes indicate ground truth spatial annotation inconsistency, green represent correct predictions from the model according to ground truth, and yellow boxes are mis-classifications (algorithm reported \textit{bicycle} and \textit{person}, but only \textit{biker} was available as label.}
  \label{fig:gtdiscrepancies}
\end{figure}

As we have shown through the experiments, there are certain inherent challenges when annotating data for deep learning and computer vision. These include lack of concise instructions, data misinterpretation due to the source data quality (low signal-to-noise ratio), cognitive degree of difficulty required for certain labeling operations, lack of annotator focus, distractions and so on. These challenges can have devastating effects if not detected early -from cost considerations to underwhelming algorithmic performance.\\
With the goal of understanding the effects and impact of data quality to algorithm performance we have conducted a series of experiments around data quality both to measure annotation quality and model performance.\\
When evaluating the quality metrics from the professional and MTurk teams, the most apparent observation we noticed can be seen in table \ref{ratings}. This table shows how different tasks can cause significantly different levels of agreement, with the average $K\mbox{-}\alpha$ easily falling within a "very good" level of agreement at $0.881$ for the VIRAT dataset, and $K\mbox{-}\alpha$ falling ~ $0.1$ short of a "good" level agreement for the Stanford Drone data set at $0.575$.\\
Table \ref{recognition} shows that \textit{Annotator 1} on the professional team, had the most disagreement from the rest of the team with an average Vitality Rating of $-0.078$. From table \ref{recognition}, we can see that the class that \textit{Annotator 1} had the most disagreement on was with the \textit{person} class, with an average Vitality Score of $-0.124$ when comparing average overall $K\mbox{-}\alpha$ for the \textit{person} class to the average $K\mbox{-}\alpha$ with \textit{Annotator 1} removed. Thus, in terms of IAA \textit{Annotator 1} showed a larger degree of disagreement with the group.\\
As far as how these observations translate to model performance, several issues arise. Comparing the models to the available labels included in the original dataset release is problematic. Treating the data with unknown annotated practices as a black box can lead to misleading conclusions about the state of the model during development. In our case, we observed that performance results with data largely unknown (\textit{gt3} dataset) were distorted as the labeling criteria varied from ours, and also, IAA was not used to filter out labels. As an example, \textit{bicycles} were originally labeled only when a person was riding them ('bikers'), not when parked. This label discrepancy can be seen in figure \ref{fig:gtdiscrepancies}, where \textit{bikes} were correctly predicted by the algorithm but they are accounted as false positives (in red) since they were not in the original set. Besides these discrepancies, there are some inconsistencies in how objects are annotated. In the same example showed, note how two pedestrians (blue boxes) next to each other have noticeable different bounding boxes around them, even though their spatial extent is similar. It seems shadows played a factor on deciding the boundaries of the objects in the original labeling exercise and those inconsistencies were not detected and filtered out. Since many of these issues are very difficult to correct, there is a need for detecting these to produce a higher quality set. Our argument is that inter annotator agreement can be a good measure to select data from annotators with high degree of agreement so that data from largely disagreeing annotators can be filtered out from the evaluating ground truth set.\\
As we can see by comparing table \ref{tab:resultsgt} against tables \ref{tab:results05gt1} and \ref{tab:results05gt2}, the performance looks very different. All versions of RetinaNet showed an increased level of performance for the inter annotator curated datasets, \textit{gt1} and \textit{gt2}, than for the original dataset \textit{gt3}. Thus, exposing the fact that results can vary substantially if we treat datasets as black boxes and ignore annotator agreement.\\
Another observation from these experiments is that for models trained with data from annotators that disagree with the majority, the model produces more candidate predictions for the same algorithm configuration (compare tables \ref{tab:results05gt1} and \ref{tab:results05gt2}). As expected, more predictions have a positive impact on recall. In contrast, when limiting the number of boxes for all models (tables \ref{tab:resultsconsgt1} and \ref{tab:resultsconsgt2}), precision and f1-score increased for a similar recall value. In both cases number of mis-classifications is substantially smaller when training a model only with data from the annotators with larger degree of agreement (\textit{i.e.} with training set \textit{drop antr 1}). This trend was observed regardless of both ground truth data selection criteria based on data quality (\textit{gt1} and \textit{gt2}).\\
Table \ref{mbox} shows that despite the restrictions placed on MTurk workers who could participate on the annotation tasks, the $K\mbox{-}\alpha$ among MTurk groups was much lower than that observed from the professional team. Furthermore, the fact that through these experiments we have seen the impact in model performance of the inclusion or removal of labeled data from annotators with larger degree of disagreement (see tables ~\ref{tab:resultsconsgt1} and \ref{tab:resultsconsgt2}), working with crowd-source teams having quality metrics throughout the exercise and constant monitoring becomes even more necessary.\\
Overall, these experiments have showed some challenges during data annotation. Including suitability of annotated data for training vs. evaluation, how to monitor data quality as annotation progresses, how to detect annotators that are failing to produce useful data, and the perils of treating original labeled data as a black box for measuring performance of custom models.\\
In addition, we have shown that annotated data quality is, indeed, a major factor when training deep learning models, as well as for measuring its performance. Not monitoring the annotation process can result in noticeable loss in algorithm precision. Finally, we have shown how the proposed extension of Krippendorff's Alpha to computer vision can be used to select training datasets that lead to higher precision models.\\
From the findings highlighter in the paper, we draw the following conclusion: The understanding of the quality of data used to train a model, as well as the clarity about the annotation process, and knowledge of the strengths and weaknesses of the ground truth data used to evaluate the models will lead to increased traceability, verification, and transparency in AI systems. Through this work we have proposed a set of tools and methodologies to facilitate AI trustworthiness assessments, which in turn will bring added value to organizations and their AI systems.\\
In the future, we want to extend this work by including more datasets and a larger sample of annotators. We want to look into how other factors (quality of instructions, mastery of skill as annotators repeat a task, annotator distraction, etc.) impact data quality and explore methodologies for early detection of annotation inconsistencies.
\section{Conclusion}
In this work we have presented a methodology for conducting annotation efforts which allows for training algorithms and measuring performance by relying on annotations with a higher levels of agreement. Often, labels included in open-source datasets are not reliable enough for industry standards. There is a need for highly efficient annotation efforts where quality is closely monitored. We have shown how inter-annotator agreement can serve as a tool to accomplish this. Therefore, it can be used by management and leadership to control the quality of data, identify individuals that need retraining, as well as identify labeled data to serve as the evaluation ground truth. 
Furthermore, we have showed how to extend and adapt Krippendorff's Alpha for computer vision. We have also shown how it can be used to consistently identify training data that leads to higher precision models.
While quality issues are inherent in any computer vision labeling effort, we identified a series of metrics that help detect and resolve issues so that task management can course correct and take action diligently. 
\bibliographystyle{aaai}
\bibliography{biblio}

\end{document}